\documentclass[11pt,a4paper]{article}
\usepackage{times,latexsym}
\usepackage{url}
\usepackage[T1]{fontenc}

\usepackage{microtype}

\usepackage[utf8]{inputenc}
\usepackage{amssymb}
\usepackage{amsmath}
\usepackage{graphicx}
\usepackage{caption}
\usepackage{tikz-dependency}
\usepackage{enumerate}
\usepackage{subcaption}

\usepackage[acceptedWithA]{tacl2021v1}
\usepackage{xspace,mfirstuc,tabulary}

\newif\iftaclinstructions
\taclinstructionsfalse 
\iftaclinstructions
\renewcommand{\confidential}{}
\renewcommand{\anonsubtext}{(No author info supplied here, for consistency with
TACL-submission anonymization requirements)}
\newcommand{\instr}
\fi

\iftaclpubformat 

\else

\fi

\title{Data-driven Parsing Evaluation for Child-Parent Interactions}

\author{
  Zoey Liu
  \\
  Boston College
  \\
  \texttt{zoey.liu@bc.edu}
  \And
  Emily Prud'hommeaux 
  \\
  Boston College
  \\
  \texttt{prudhome@bc.edu}
}

\date{}

\begin{document}
\maketitle
\begin{abstract}

We present a syntactic dependency treebank for naturalistic child and child-directed speech in English~\cite{macwhinney2000childes}. Our annotations largely followed the guidelines of the Universal Dependencies project (UD~\citep{11234/1-4758}), with detailed extensions to lexical/syntactic structures unique to conversational speech (in opposition to written texts). Compared to existing UD-style spoken treebanks as well as other dependency corpora of child-parent interactions specifically, our dataset is of (much) larger size ($N$ of utterances = 44,744; $N$ of words = 233, 907) and contains speech from a total of 10 children covering a wide age range (18-66 months).
With this dataset, we ask:
(1) How well would state-of-the-art dependency parsers, tailored for the written domain, perform for speech of different interlocutors in spontaneous conversations?
(2) What is the relationship between parser performance and the developmental stage of the child?
To address these questions, in ongoing work, we are conducting thorough dependency parser evaluations using both graph-based and transition-based parsers with different hyperparameterization, trained from three different types of out-of-domain written texts: news, tweets, and learner data.
Our dataset and code are available at \url{https://github.com/zoeyliu18/childes_parsing}.


\end{abstract}


\section{Introduction}

Research on syntactic dependency parsing has undoubtedly enjoyed tremendous progress with the continuous development of the Universal Dependencies project~\citep{11234/1-4758} (hereafter UD).
That said, of the 228 treebanks in the latest version of UD (v2.10 at the time of writing), only 12 are treebanks of fully spoken data (see also~\citet{dobrovoljc2022}), while the rest focus on different genres within the written domain (e.g. news, wikipedia texts).
This means that most (if not all) off-the-shelf dependency parsers that are considered state-of-the-art are oriented towards written texts, rather than tailored specifically for spontaneous speech.
Therefore a natural question arises: how well would parsing systems developed for written data perform when it comes to spontaneous speech?

Over the past decade, there have been efforts devoted into dependency parsing for the spoken domain, especially for (a subset of) the Switchboard corpus~\citep{godfrey1992switchboard}, which contains transcripts of telephone conversations in English; while some focused on parsing the full subset~\citep{yoshikawa-etal-2016-joint,rasooli-tetreault-2013-joint,miller-2008-unified-syntac}, others attended to specific phenomena common in spoken data, such as speech repairment~\citep{miller-2009-improv-syntac,miller-2009-improved}.
In addition to English, dependency treebanks have been developed for speech of other languages, including but not limited to French~\cite{gerdes-2009-speak,bazillon-2012-syntac}, Czech~\citep{mikulova-2017-pdtsc}, Russian~\citep{kovriguina-2018-multil-annot}, Japanese~\citep{10.1093/ietisy/e88-d.3.545} and Mandarin Chinese~\citep{he-2018-depen-parser}. These treebanks, including Switchboard, however, were not always built on the basis of the UD guidelines; the annotations and trained parsers are not always publicly available, making it less straightforward to perform parsing evaluation with the data, especially given the fact that the majority of dependency parsers are specific to UD-formatted treebanks.

This paper presents a wide-coverage dataset of spontaneous child-parent interactions~\citep{macwhinney2000childes}, annotated with syntactic dependencies largely following the UD standards. 
We illustrate careful annotation guidelines that hopefully would be useful to other developments of spoken dependency treebanks.
Compared to previous studies, our work goes beyond in several respects (see Section~\ref{related}). 
First, compared to most of the other spoken dependency treebanks that contain telephone conversations~\citep{bechet-2014-adapt}, interactions between adults~\citep{dobrovoljc-2018-er,dobrovoljc-2016-univer-depen}, or user-generated content~\citep{davidson-2019-depen-parsin}, our dataset attends to child and child-directed speech.

Second, in contrast to other spoken treebanks in the UD project, our dataset as a whole is of considerable size: there are 26,098 utterances ($N$ of words = 116,428) for child speech, and 18,646 utterances ($N$ of words = 117,479) for parent speech.

Third, while there are some dependency corpora of child-parent interactions in English~\citep{sagae-2010-morph-annot}, Japanese~\citep{childes-japanese}, and Hebrew~\citep{gretz-2013-parsin-hebrew}, they only include data from one or two children, whereas  we provide annotations for the speech of 10 children across a wider age range, therefore covering in more details lexical/syntactic phenomena that are more unique to child speech (e.g., repetition, speech disfluency) and spoken language more broadly.

With this dataset, in ongoing work, we ask two additional questions:
(1) How do state-of-the-art dependency parsing architectures trained on out-of-domain data perform when it comes to naturalistic speech of separate interlocutors?
To address this question, we evaluate parsers trained on three genres within the written domain: news texts, tweets and learner data, all in English.
(2) What is the relationship, if any, between parser performance and the developmental stage of the child? In a way one might foresee a positive correlation between the two, with the expectation that as the child continues to develop their language skills, they would utilize more and more cohesive syntactic structures, instead of, for example, produce speech with unintelligible speech or word omissions. On the other hand, it is possible that parser performance would increase as the child reaches a certain developmental stage, then start decreasing, since the child might start articulating sentences with more complex or expressive syntactic structures, which thereby are potentially harder to analyze.

\begin{table*}[h!]
\footnotesize
\centering
\begin{tabular}{c|c|c|c|c|c|c|c|c}
\hline
\textbf{Child} & \textbf{18-24} & \textbf{24-30} & \textbf{30-36} & \textbf{36-42} & \textbf{42-48} & \textbf{48-54} & \textbf{54-60} & \textbf{60-66}  \\\hline
Violet & 1,730 & 1,855 & 1,889 & 1,886 & 1,894 &   &   \\
Parent & 2,007 & 1,995 & 2,025 & 2,044 & 2,030 &   &   \\\hline
Naima & 1,748 & 1,752 & 1,814 & 1,902 & 1,943 &   &   \\
Parent & 2,019 & 2,007 & 2,005 & 2,032 & 2,019 &   &   \\\hline
Thomas &   & 1,969 & 1,965 & 1,972 & 1,984 & 2,009 & 1,992 \\
Parent &   & 2,013 & 2,035 & 2,050 & 2,016 & 2,025 & 2,010 \\\hline
Adam &   & 2,012 & 2,004 & 2,016 & 2,021 & 2,012 & 2,026 & 2,025 \\
Parent &   & 2,012 & 2,009 & 2,020 & 2,008 & 2,015 & 2,007 & 1,234 \\\hline
Lily & 2,004 & 1,788 & 1,908 & 1,981 & 1,957 & 1,161 &   \\
Parent & 1,996 & 2,016 & 1,999 & 2,038 & 2,006 & 2,000 &   \\\hline
Laura & 1,930 & 1,925 & 1,955 & 1,912 & 1,956 & 1,943 & 1,954 \\
Parent & 2,022 & 2,019 & 2,013 & 1,991 & 2,023 & 1,984 & 2,015 \\\hline
Roman &   & 1,980 & 1,975 & 1,990 & 2,004 & 1,999 & 2,009 \\
Parent &   & 1,712 & 2,017 & 2,012 & 2,020 & 2,013 & 2,027 \\\hline
Emma &   &   & 2,007 & 2,014 & 2,029 & 2,025 &   \\
Parent &   &   & 2,020 & 2,038 & 2,015 & 1,455 &   \\\hline
Sarah &   & 2,014 & 2,008 & 2,011 & 2,007 & 2,020 & 2,015 & 2,028 \\
Parent &   & 2,019 & 2,043 & 2,037 & 2,036 & 2,026 & 2,032 & 2,049 \\\hline
Abe &   & 2,007 & 2,007 & 2,022 & 2,020 & 2,036 & 2,021 & 1,386 \\
Parent &   & 2,028 & 2,040 & 2,025 & 2,020 & 2,017 & 2,019 \\\hline
\end{tabular}
\caption{$N$ of words for child and parent speech at different age ranges of the children.}
\label{tab:descriptive_stats}
\end{table*}

\section{Related Work}
\label{related}

Earlier work on dependency parsing of child-parent conversations~\citep{sagae-etal-2001-parsing,sagae-etal-2004-adding} focused on the Eve corpus~\citep{brown1973first} from the CHILDES project~\citep{macwhinney2000childes}, though the annotations did not follow those of UD. 
Subsequent research extended the annotation guidelines for the Eve corpus~\citep{sagae-2010-morph-annot} to child and child-directed speech in Japanese~\citep{childes-japanese} and Hebrew~\citep{gretz-2013-parsin-hebrew}.

We note two studies that carried out UD-style dependency parse annotations for child and/or child-directed speech..
~\citet{liu-prudhommeaux-2021-dependency} took a semi-automatic approach to convert part of the existing dependency parses from the Eve corpus~\citep{brown1973first} to UD standards; specifically, they focused on child and parent utterances when the child is within 18-27 months old. Concurrent work by ~\citet{szubert2021cross} annotated dependency parses and semantic logical forms for two languages: English (a large portion from the Adam corpus~\citep{brown1973first}) and Hebrew (The Hagar corpus~\citep{BERMAN+1990+1135+1166}), although they only looked at child-directed speech.

\section{Meet the Data} 

For dataset construction, we borrowed transcripts of English naturalistic  parent-child interactions from the \texttt{childes-db} interface~\citep{sanchez2019childes}, which contains data from the CHILDES~\citep{macwhinney2000childes} database. As we are interested in how parser evaluation results change at different developmental stages of a child, we used \textit{age} as a proxy for developmental stage and set 6-month as one age bin.
For every unique child from each corpus of English, we calculated the total number of words produced by the child ($N\_child$) and by the parent(s) ($N\_parent$; excluding data from other care-givers) respectively within each age bin of the child.
From each age bin, for both child and parent speech, we randomly sampled a number of utterances (mostly without replacement) that amounted to approximately 2,000 words; the criteria was relaxed somewhat in order to include data across a wide range of age bins.
This resulted in spoken data of ten children from 6 corpora (Table~\ref{tab:descriptive_stats}).
In what follows, we briefly describe each of the selected corpora.

\noindent \textbf{Kuczaj}~\citep{kuczaj1977acquisition} The Kuczaj corpus includes speech from diary study of the child, Abe; each original recording lasts $\sim$30 minutes.

\noindent \textbf{Brown}~\citep{brown1973first} From the Brown corpus we included naturalistic speech from Adam and Sarah.

\noindent \textbf{Thomas}~\citep{lieven2009two} The Thomas corpus contains spoken interactions between the child, Thomas, and primarily his mother at their house; each initial audio recording is about an-hour long.

\noindent \textbf{Weist}~\citep{weist2008autobiographical} The data of Emma and Roman came from the Weist corpus, which includes caregiver-child interactions recorded in either a laboratory setting or in their own homes twice a month for around 30 minutes.

\noindent \textbf{Braunwald} We used recorded longitudinal speech of Laura when interacting with different interlocutors; the speech was produced under naturalistic environments.

\noindent \textbf{Providence} We took data of three children, Naima, Lily and Violet, from the Providence corpus~\citep{demuth2006word}; the data contains longitudinal spontaneous interactions at home between the children and mostly their mothers.


\section{Annotation process}

Our annotations largely followed those of UD~\citep{11234/1-4758}.
Annotator A, who has advanced  training in dependency syntax, initially annotated data of age 18-24 months and 24-30 months from Abe and Sarah, as a way to take notes of any domain-specific phenomenon or cases that might not be straightforward to annotate (see Section~\ref{guidelines}). 
These guidelines were discussed with annotator B and modified if needed.
Then given each age bin of every child, the two annotators annotated 10\% of data from both child and parent speech. 
We calculated agreement scores using Cohen's Kappa~\cite{artstein2008inter}.
The overall agreement score taking into account all syntactic head and dependency relation annotations of all data is 0.97; the average agreement score across each dependency parse is 0.96. (We also computed agreement scores for each child and the results are around 0.97). Final annotations of all data were performed and checked by annotator A.

\section{Annotation guidelines}
\label{guidelines}

Here we describe in details our approach to transcription orthography, tokenization and dependency annotations for syntactic constructions that are unique or more common in child speech and spoken data more broadly.

\subsection{Orthography and tokenization}

Regarding orthography of the transcripts, we made four decisions, all of which are on the basis of a principle that we call ``annotate what is actually there".
First, we did not perform any orthographic normalization of most intelligible words in the speech (e.g., \textit{she wana eat}); in other words, these words stayed true to their original forms taken from CHILDES.
That said, the tokenizations of certain cases were updated following UD. These cases include: (1) possessives (e.g., \textit{Daddy's} $\rightarrow$ \textit{Daddy 's}); (2) shortened copulas (e.g., \textit{I'm eating} $\rightarrow$ \textit{I 'm eating}); (3) combined conjunctives (e.g., \textit{in\_spite\_of} $\rightarrow$ \textit{in spite of} ); (4) combined adverbs (e.g., \textit{as\_well} $\rightarrow$ \textit{as well}); (5) negation (e.g., \textit{don't} $\rightarrow$ \textit{do n't}); (6) other informal contraction (e.g., \textit{gonna} $\rightarrow$ \textit{gon na}); (7) childish expressions (e.g., \textit{poo\_poo}, \textit{choo\_choo})

Second, unintelligible speech words (e.g., \textit{xxx}, \textit{yyy}) were removed as it is hard to tell whether the words existed in the first place and whether they have any syntactic roles in the utterances.

Third, we kept the initial capitalization in the transcripts, since most of the times only proper names or words such as \textit{Mommy} were capitalized.

Lastly, we omitted all punctuations with the exceptions of apostrophe (to abide with UD's standards), since punctuations tend to not be explicitly articulated in actual spontaneous speech; they are instead added manually during the transcription process based on judgments by the annotators.
Admittedly, under certain circumstances it is relatively easy to make choices of punctuations; for instance, one could use a question mark at the end of an utterance with a raising pitch contour.
That said, deciding which punctuation to use is not always easy and could be quite time-consuming; for example, for spoken data, there could always be the problem of setting utterance boundary (see Section~\ref{utterance_boundary}), and accordingly, whether to insert, say, a period or a semi-colon is not as simple as one might think. Therefore the process of adding punctuations in transcripts is quite subjective itself.

\subsection{(Vague) utterance boundary}
\label{utterance_boundary}
\begin{figure}[h!]
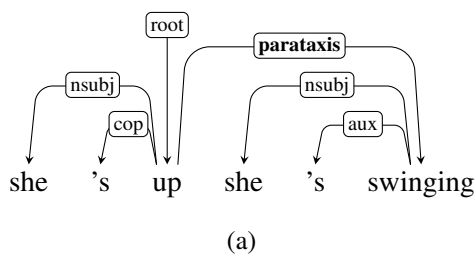

\centering
\begin{minipage}{0.5\textwidth}
  \centering
 \begin{dependency}
   \begin{deptext}[column sep=1em]
she \& 's \& up \& she \& 's \& swinging  \\
   \end{deptext}
   \deproot{3}{root}
   \depedge{3}{1}{nsubj}
   \depedge{3}{6}{\textbf{parataxis}}
   \depedge{3}{2}{cop}
   \depedge{6}{4}{nsubj}
   \depedge{6}{5}{aux}
\end{dependency}
\subcaption[first caption.]{}\label{fig:utt_boundary_a}
\end{minipage}%
\caption{Examples of \textit{(vague) utterance boundary}.} \label{fig:utt_boundary}
\end{figure}

Each instance in the dataset was originally treated as one utterance in CHILDES, therefore we tried to annotate them as one sentence most of the times. That said, the initial utterance boundaries are not always adequate. This means that one extracted instance can be considered as having ``side-by-side" sentences (Figure~\ref{fig:utt_boundary_a})
\footnote{Examples presented in this paper are often modified from the original utterances for ease of presentation.}
; we abided by the instructions of UD and annotated the first sentence to be the root; then later sentences in the instance were treated as \textit{parataxis} of the root.

\subsection{Creative lexical usage}

\begin{figure}[h!]
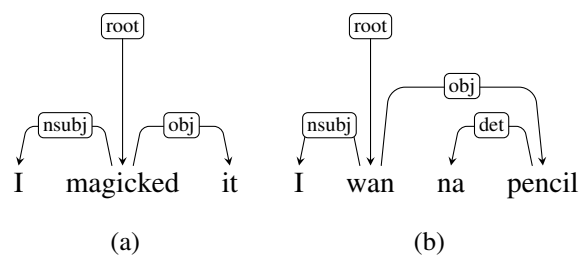

\centering
\begin{minipage}{0.25\textwidth}
  \centering
 \begin{dependency}
   \begin{deptext}[column sep=1em]
I \& magicked \& it  \\
   \end{deptext}
   \deproot{2}{root}
   \depedge{2}{1}{nsubj}
   \depedge{2}{3}{obj}
\end{dependency}
\subcaption[first caption.]{}\label{fig:creative_a}
\end{minipage}%
\begin{minipage}{0.25\textwidth}
  \centering
 \begin{dependency}
   \begin{deptext}[column sep=1em]
I \& wan \& na \& pencil  \\
   \end{deptext}
   \deproot{2}{root}
   \depedge{2}{1}{nsubj}
   \depedge{2}{4}{obj}
   \depedge{4}{3}{det}
\end{dependency}
\subcaption[first caption.]{}\label{fig:creative_b}
\end{minipage}%
\caption{Examples of \textit{creative lexical usage}.} 
\end{figure}

It is common that children make certain lexical choices that do not necessarily follow the standards of parent speech or (formal) written data (e.g., \textit{mine pillow}; \textit{my mummy telephones me}). On the other hand, these cases in a way reflect children's world, in the sense that they capture children's own understanding of these words and their lexical (and syntactic) development at different stages. Therefore  here we refer to such cases as \textit{creative usage}; for each case, we analyzed their syntactic usage given the remaining structure of the sentence~\cite{lee-etal-2017-towards,santorini1990part}, then assigned dependency parses accordingly.
For example, in Figure~\ref{fig:creative_a}, the word \textit{magicked} is creatively used as a verb that links the subject \textit{I} and object \textit{it}.

For some instances, it is relatively difficult to decide whether the utterance contains the child's creative usage of some lexical items, or potential transcription errors induced by annotators. Given that transcribing spoken data manually requires large amounts of time and energy, it is not against expectations that the resulting transcriptions might have errors.
For instance, with Figure~\ref{fig:creative_b}, it is not exactly clear whether the child really said \textit{I wan na pencil}, or the transcription should have been \textit{I want a pencil}. Our approach is that we compared how often each alternative occurs in our dataset, then made the final decision. Therefore between the two alternatives above, we chose to annotate \textit{na} as the determiner of \textit{pencil}.

\subsection{\textit{Possible} lexical omission}

As we are taking a data-driven approach, we tried to perform annotations based on how words or phrases are used within an utterance; this means that we avoided assuming potential word omissions as much as possible.
We assigned dependency structures to an utterance if a reasonable parse (given the context) could be derived without the assumption that certain words are missing.

\begin{figure}[h!]
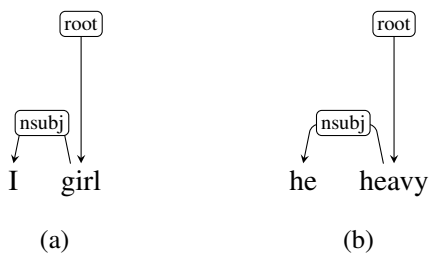

\centering
\begin{minipage}{0.25\textwidth}
  \centering
 \begin{dependency}
   \begin{deptext}[column sep=1em]
I \& girl  \\
   \end{deptext}
   \deproot{2}{root}
   \depedge{2}{1}{nsubj}
\end{dependency}
\subcaption[first caption.]{}\label{fig:omission_a}
\end{minipage}%
\begin{minipage}{0.25\textwidth}
  \centering
 \begin{dependency}
   \begin{deptext}[column sep=1em]
he \& heavy  \\
   \end{deptext}
   \deproot{2}{root}
   \depedge{2}{1}{nsubj}
\end{dependency}
\subcaption[first caption.]{}\label{fig:omission_b}
\end{minipage}%
\caption{Examples of \textit{possible lexical omission}: omission of copula.} 
\end{figure}

\begin{figure*}[h!]
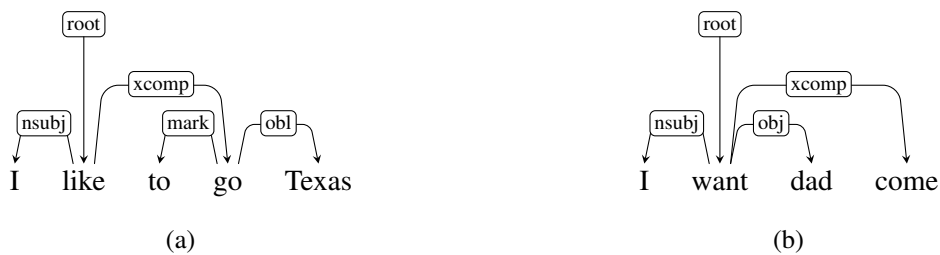

\centering
\begin{minipage}{0.5\textwidth}
  \centering
 \begin{dependency}
   \begin{deptext}[column sep=1em]
I \& like \& to \& go \& Texas  \\
   \end{deptext}
   \deproot{2}{root}
   \depedge{2}{1}{nsubj}
   \depedge{2}{4}{xcomp}
   \depedge{4}{3}{mark}
   \depedge{4}{5}{obl}
\end{dependency}
\subcaption[first caption.]{}\label{fig:omission_c}
\end{minipage}%
\begin{minipage}{0.5\textwidth}
  \centering
 \begin{dependency}
   \begin{deptext}[column sep=1em]
I \& want \& dad \& come  \\
   \end{deptext}
   \deproot{2}{root}
   \depedge{2}{1}{nsubj}
   \depedge{2}{3}{obj}
   \depedge{2}{4}{xcomp}
\end{dependency}
\subcaption[first caption.]{}\label{fig:omission_d}
\end{minipage}%
\caption{Examples of \textit{possible lexical omission}: omission of adposition.} 
\end{figure*}

In other cases, we deemed the utterance as having lexical omission if the omitted word could be \textit{automatically retrievable}; this way other researchers could formulate a different analysis for the utterance as they see fit.
In our annotations, we considered two types of lexical omissions.
The first type is copula omission, where the syntactic head of the copula is mostly a noun (Figure~\ref{fig:omission_a}), an adjective ((Figure~\ref{fig:omission_b}), or an adposition ((Figure~\ref{fig:omission_c}); other times we assumed that a copula is omitted if the utterance can be interpreted as an expletive structure (e.g., \textit{there book}, with \textit{there} as the \textit{expl} dependent of \textit{book}).
The second type is adposition omission, mostly the adposition that is the function head of an oblique phrase (Figure~\ref{fig:omission_c}) or the infinitival-\textit{to} in a complement clause (Figure~\ref{fig:omission_d}).

\subsection{Nominal phrases}

\begin{figure*}[h!]
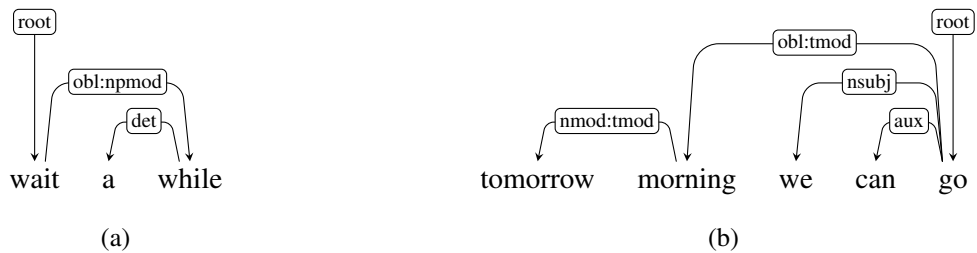

\centering
\begin{minipage}{0.5\textwidth}
  \centering
 \begin{dependency}
   \begin{deptext}[column sep=1em]
wait \& a \& while \\
   \end{deptext}
   \deproot{1}{root}
   \depedge{1}{3}{obl:npmod}
   \depedge{3}{2}{det}
\end{dependency}
\subcaption[first caption.]{}\label{fig:time_a}
\end{minipage}%
\begin{minipage}{0.5\textwidth}
  \centering
 \begin{dependency}
   \begin{deptext}[column sep=1em]
tomorrow \& morning \& we \& can \& go \\
   \end{deptext}
   \deproot{5}{root}
   \depedge{5}{2}{obl:tmod}
   \depedge{5}{3}{nsubj}
   \depedge{5}{4}{aux}
   \depedge{2}{1}{nmod:tmod}
\end{dependency}
\subcaption[first caption.]{}\label{fig:time_b}
\end{minipage}%
\caption{Examples of \textit{nominal phrases}.} 
\end{figure*}

For certain nominal phrases that serve as adverbial modifiers in a given utterance (e.g., Figure~\ref{fig:time_a}), and/or express time and dates, we tried to annotate them more carefully using subtypes of specific dependency relations (e.g., \textit{nmod:tmod} or \textit{obl:tmod})~\citep{schneider-zeldes-2021-mischievous}. For example, in Figure~\ref{fig:time_b}, depending on their respective role, \textit{morning} should be an oblique phrase of \textit{go} whereas \textit{tomorrow} modifies \textit{morning}).

\subsection{Ambiguity}

The syntactic structure of a sentence can be ambiguous when looking at the sentence by itself. Therefore we took into account the surrounding context of an utterance when performing annotations.
In some cases, contexts could be helpful; for example, in (1), \textit{like} can be treated as the verb of the sentence, rather than an adposition.
\begin{enumerate}
    \item[(1)] \textit{Parent: do you like this}
    \item[] \textit{Child: like this}
\end{enumerate}
In other (rare) cases, context might not be quite useful; for instance, in (2), it is not clear whether \textit{rain} should be a verb and the relation between the two words is \textit{obl:tmod}, or a noun and the dependency relation is \textit{nmod:tmod}. For these examples we opted for the simpler analysis given characteristics of child speech and treated \textit{rain} as a noun.
\begin{enumerate}
    \item[(2)] \textit{Parent: eat your soup please}
    \item[] \textit{Child: rain tonight}
\end{enumerate}

Another source of ambiguity comes from whether to treat proper names or words like \textit{mommy} and \textit{daddy} as \textit{vocative} or not, e.g., \textit{Momma try it}. For these cases, we decided to consider them as \textit{vocative} if this interpretation is reasonable, since subject omission is common in early child speech~\citep{hughes2006discourse}.

\subsection{Non-canonical word order}

\begin{figure}[h!]
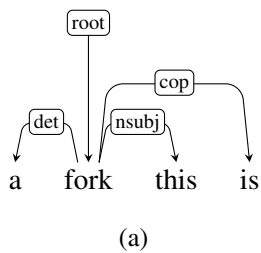

\centering
\begin{minipage}{0.5\textwidth}
  \centering
 \begin{dependency}
   \begin{deptext}[column sep=1em]
a \& fork \& this \& is  \\
   \end{deptext}
   \deproot{2}{root}
   \depedge{2}{1}{det}
   \depedge{2}{3}{nsubj}
   \depedge{2}{4}{cop}
\end{dependency}
\subcaption[first caption.]{}\label{fig:order_b}
\end{minipage}%
\caption{Examples of non-canonical word order.} 
\end{figure}

Often times in child speech or spoken data more broadly, an utterance does not have the canonical word order that is presumed in (formal) written data. In our dataset, cases as such usually involve post-posed subject and (copula) verb (Figure~\ref{fig:order_b}). For these cases, we assigned dependency relations based on the syntactic function of a word/phrase which is not constrained by their relative orderings.

\subsection{Speech repairment}

\begin{figure*}[h!]
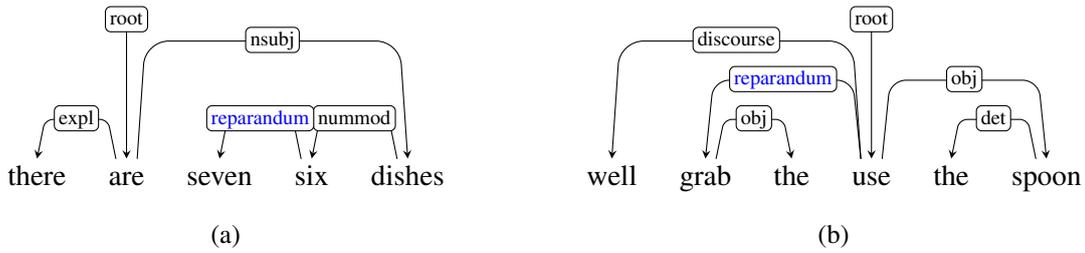

\centering
\begin{minipage}{0.5\textwidth}
  \centering
 \begin{dependency}
   \begin{deptext}[column sep=1em]
there \& are \& seven \& six \& dishes \\
   \end{deptext}
   \deproot{2}{root}
   \depedge{2}{1}{expl}
   \depedge{2}{5}{nsubj}
   \depedge{5}{4}{nummod}
   \depedge{4}{3}{\textcolor{blue}{reparandum}}
\end{dependency}
\subcaption[first caption.]{}\label{fig:repairment_a}
\end{minipage}%
\begin{minipage}{0.5\textwidth}
  \centering
\begin{dependency}
   \begin{deptext}[column sep=1em]
well \& grab \& the \& use \& the \& spoon  \\
   \end{deptext}
   \deproot{4}{root}
   \depedge{4}{1}{discourse}
   \depedge{2}{3}{obj}
   \depedge{4}{2}{\textcolor{blue}{reparandum}}
   \depedge{4}{6}{obj}
   \depedge{6}{5}{det}
\end{dependency}
\subcaption[second caption.]{}\label{fig:repairment_b}
\end{minipage}%
\caption{Examples of  \textit{repairment}; the dependency relation for speech repair in each example is in \textcolor{blue}{blue}.} \label{fig:repairment}
\end{figure*}

For speech repair, which captures one type of disfluency~\citep{FERREIRA2004231}, we used \textit{reparandum} in the same way as suggested by the UD guidelines, that is, the speech repair is the syntactic head of the subtree that constitues the disfluent speech (e.g., \textit{seven} in Figure~\ref{fig:repairment_a}).
If the disfluent speech contains discourse fillers or editing terms such as \textit{uh} or \textit{um}, these elements are annotated to be the syntactic dependents of the repair with the relation \textit{discourse} (also to avoid unnecessary crossing dependencies).
In some more complicated cases, the disfluency subtrees are word fragments that do not form a whole coherent phrase together (e.g., \textit{grab the} in Figure~\ref{fig:repairment_b}); for these cases, we used the principle of promotion to analyze elements within the subtree structure of the disfluency if needed. For example, with Figure~\ref{fig:repairment_b}, the word \textit{grab} is most likely to be the head within the disfluency subtree, therefore we promoted the following \textit{the} to be the \textit{object} of \textit{grab}, then analyzed the dependency relations of the residual structures in the instance.

To separate repairment from speech restart or abandonment (Section~\ref{restart}), we only categorized an utterance as having speech repairment if the repairment occurs in-between the utterance / is sentence-medial (as opposed to the beginning of the sentence).

\subsection{Speech restart}
\label{restart}

Another type of disfluency is speech restart.
We generally considered an instance as having speech restart or abandonment if the abandoned elements occur at the beginning of the instance and do not form a coherent phrase together; what's more, the abandoned elements need to be different from the speech restart. For these cases, given that speech restart falls broadly under the umbrella of disfluency, and in order to distinguish restart from repairment above, we extended \textit{reparandum} with a new dependency relation subtype: \textit{reparandum:restart}
 to connect the abandoned elements as the dependents of the speech restart. This way the dependency relation will also go ``right-to-left"~\citep{dobrovoljc2022}, following the usage of \textit{reparandum}.

\begin{figure}[h!]
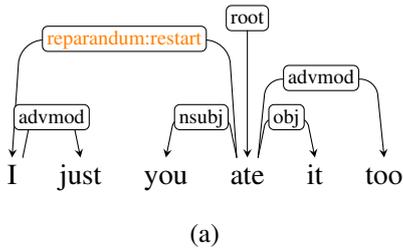

\centering
\begin{minipage}{0.5\textwidth}
  \centering
 \begin{dependency}
   \begin{deptext}[column sep=1em]
I \& just \& you \& ate \& it \& too  \\
   \end{deptext}
   \deproot{4}{root}
   \depedge{4}{1}{\textcolor{orange}{reparandum:restart}}
   \depedge{4}{3}{nsubj}
   \depedge{4}{5}{obj}
   \depedge{4}{6}{advmod}
   \depedge{1}{2}{advmod}
\end{dependency}
\subcaption[first caption.]{}\label{fig:restart_a}
\end{minipage}%
\caption{Examples of \textit{restart}; the dependency relation between repeated elements in each example is in \textcolor{orange}{orange}.} \label{fig:restart}
\end{figure}

\subsection{Repetition}
\label{repetition}

\begin{figure*}[h!]
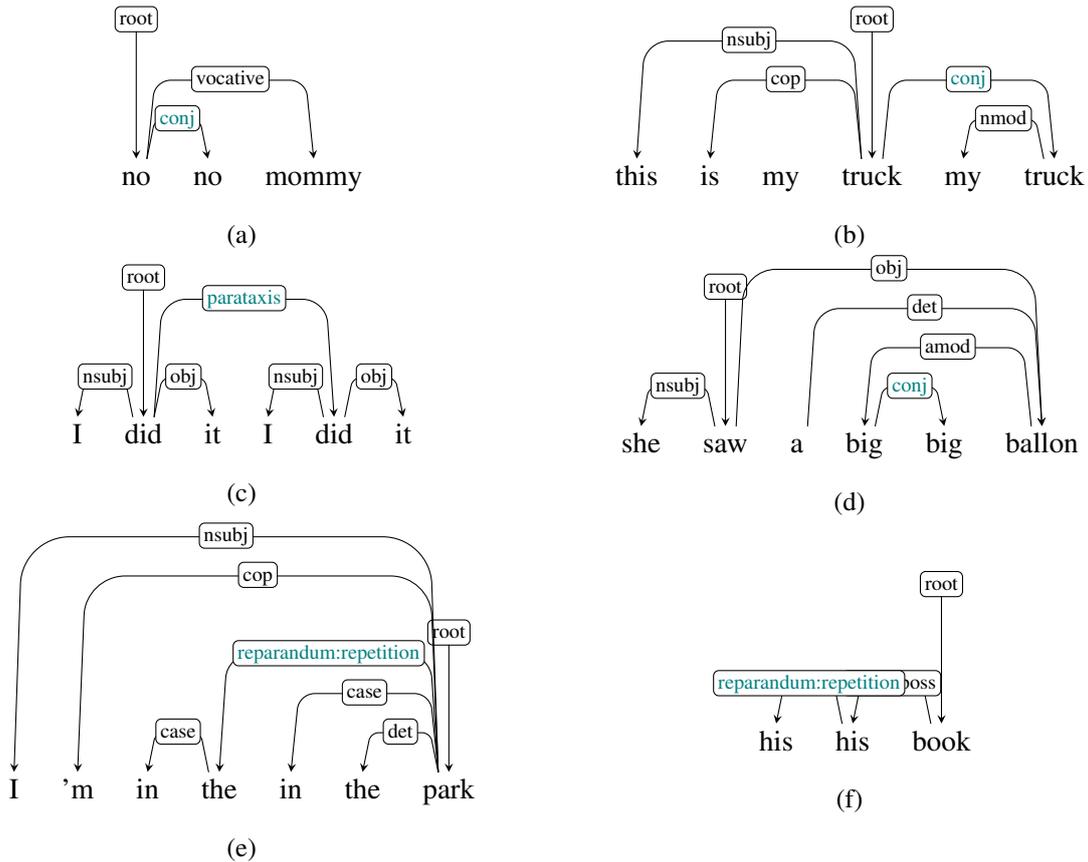

\centering
\begin{minipage}{0.5\textwidth}
  \centering
 \begin{dependency}
   \begin{deptext}[column sep=1em]
no \& no \& mommy  \\
   \end{deptext}
   \deproot{1}{root}
   \depedge{1}{2}{\textcolor{teal}{conj}}
   \depedge{1}{3}{vocative}
\end{dependency}
\subcaption[first caption.]{}\label{fig:repetition_a}
\end{minipage}%
\begin{minipage}{0.5\textwidth}
  \centering
\begin{dependency}
   \begin{deptext}[column sep=1em]
this \& is \& my \& truck \& my \& truck  \\
   \end{deptext}
   \deproot{4}{root}
   \depedge{4}{1}{nsubj}
   \depedge{4}{2}{cop}
   \depedge{6}{5}{nmod}
   \depedge{4}{6}{\textcolor{teal}{conj}}
\end{dependency}
\subcaption[second caption.]{}\label{fig:repetition_b}
\end{minipage}%
\\
\begin{minipage}{0.5\textwidth}
  \centering
\begin{dependency}
   \begin{deptext}[column sep=1em]
I \& did \& it \& I \& did \& it  \\
   \end{deptext}
   \deproot{2}{root}
   \depedge{2}{1}{nsubj}
   \depedge{2}{3}{obj}
   \depedge{2}{5}{\textcolor{teal}{parataxis}}
   \depedge{5}{4}{nsubj}
   \depedge{5}{6}{obj}
\end{dependency}
\subcaption[third caption.]{}\label{fig:repetition_c}
\end{minipage}%
\begin{minipage}{0.5\textwidth}
  \centering
\begin{dependency}
   \begin{deptext}[column sep=1em]
she \& saw \& a \& big \& big \& ballon  \\
   \end{deptext}
   \deproot{2}{root}
   \depedge{2}{1}{nsubj}
   \depedge{2}{6}{obj}
   \depedge{3}{6}{det}
   \depedge{6}{4}{amod}
   \depedge{4}{5}{\textcolor{teal}{conj}}
\end{dependency}
\subcaption[fourth caption.]{}\label{fig:repetition_d}
\end{minipage}%
\\
\begin{minipage}{0.5\textwidth}
  \centering
 \begin{dependency}
   \begin{deptext}[column sep=1em]
I \& 'm \& in \& the \& in \& the \& park  \\
   \end{deptext}
   \deproot{7}{root}
   \depedge{7}{1}{nsubj}
   \depedge{7}{2}{cop}
   \depedge{4}{3}{case}
   \depedge{7}{4}{\textcolor{teal}{reparandum:repetition}}
   \depedge{7}{5}{case}
   \depedge{7}{6}{det}
\end{dependency}
\subcaption[fifth caption.]{}\label{fig:repetition_e}
\end{minipage}%
\begin{minipage}{0.5\textwidth}
  \centering
 \begin{dependency}
   \begin{deptext}[column sep=1em]
his \& his \& book  \\
   \end{deptext}
   \deproot{3}{root}
   \depedge{3}{2}{nmod:poss}
   \depedge{2}{1}{\textcolor{teal}{reparandum:repetition}}
\end{dependency}
\subcaption[sixth caption.]{}\label{fig:repetition_f}
\end{minipage}%
\caption{Examples of \textit{repetition}; the dependency relation between repeated elements in each example is in \textcolor{teal}{teal}.} \label{fig:repetition}
\end{figure*}

Overall we identify three major kinds of repetitions. For the first type, an utterance consists of repetitions of the same dependency subtree and the repeated subtree is a coherent phrase by itself. Examples include cases such as discursive repetition (e.g., \textit{no no mommy}; Figure~\ref{fig:repetition_a}), onomatopoeia (e.g., \textit{honk honk}), or repetition of other kinds of word or phrase (e.g., \textit{this is my truck my truck}; Figure~\ref{fig:repetition_b}). For these cases, we treated the first appearance of the repeated subtree as the syntactic head with the following repetitions as the dependents connected with the relation \textit{conj}. 
A special case is when the instance repeats a full sentence (or just containing a verbal phrase?), e.g., \textit{I did it I did it} (Figure~\ref{fig:repetition_c}); for these examples we used \textit{parataxis} to adhere to the annotations of side-by-side sentences noted by UD.

For the second type, repetition is used to emphasize the characteristics of certain objects or conditions (or serves as an intensifier~\citep{szubert2021cross}); in these cases the repeated element is usually a single word with a POS tag of adjective or adverb. These cases were annotated similarly as those from the first type above (Figure~\ref{fig:repetition_d}).

The third type of repetition pertains disfluency. Whether to interpret an instance as disfluent repetitions or not could be challenging when the instance does not have a corresponding audio, where the prosody could help guide the interpretation. Therefore to distinguish the two types mentioned before, we considered an instance as having disfluent repetitions if the repetition appears at the beginning or in-between a sentence; in addition, the repeated element has to be either word fragments that do not form a whole coherent phrase together when taking the sentential context into account (Figure~\ref{fig:repetition_e}), or a single word and its POS tag is neither an adjective or an adverb (Figure~\ref{fig:repetition_f}).
For these cases, in order to not go against the fact that \textit{conj} is usually applied left-to-right (i.e., syntactic head precedes its dependents),  we again extended the usage of \textit{reparandum} and applied a new dependency relation subtype: \textit{reparandum:repetition} to describe repetition in disfluency; speech repairment, restart and disfluent repetition hence would also be easily distinguishable in an automatic fashion.

\subsection{Other structures}

The last three types of structures to be mentioned are serial verb construction (SVC), tag question and \textit{unintelligible} structure where we could not decide on a clear (or sometimes any) interpretation.
For SVC, we followed~\citet{szubert2021cross}; for an utterance such as \textit{he came see me play}, \textit{see} was treated as the dependent of \textit{came} and the relation is \textit{compound:svc}.

Tag questions are mostly used in parent speech (e.g., \textit{you like that book do n't you}); for these examples we abided by the UD annotations and used \textit{parataxis} to connect the tag question to the main clause of the utterance.

Lastly, for utterances without a clear syntactic structure, we used \textit{dep} to connect the individual word fragments. Given that UD advises to avoid applying this relation as much as possible, we restricted its usage to mostly utterances produced in early developmental stages of the child (e.g., 18-24 months), where the utterances are composed of usually two to three words in total.




\section{Experiments}


\subsection{Meet the parsers}
\label{parsers}

We experimented with two graph-based parsers and one transition-based parser using their default parameters: (1) \textbf{Diaparser}~\citep{attardi-etal-2021-biaffine}: A graph-based biaffine parser model~\citep{DBLP:conf/iclr/DozatM17} which uses a language model (LM) with three BiLSTM layers as encoder. 
(2)  \textbf{MaChamp}~\citep{van-der-goot-etal-2021-massive}: A multi-task learning toolkit which includes the option of dependency parsing; the parser is also a graph-based biaffine parser~\citep{DBLP:conf/iclr/DozatM17}.
To increase the variety of parser hyperparameterization, for both Diaparser and MaChamp, we explored three LMs as encoders: \texttt{bert-base-cased} (\texttt{bert})~\citep{DBLP:journals/corr/abs-1810-04805}, \texttt{roberta-base} (\texttt{roberta})~\citep{DBLP:journals/corr/abs-1907-11692}, and
\texttt{twitter-roberta-base} (\texttt{twitter})~\citep{barbieri-etal-2020-tweeteval}. 
(3) \textbf{UUParser}~\citep{delhoneux17arc,smith-etal-2018-82}: A greedy transition-based parser that extends the model from~\citet{10.1162/tacl_a_00101} via allowing predictions to have non-projective dependency trees~\citep{nivre-2009-non}.


\subsection{Out-of-domain data}

\begin{table}[h!]
\footnotesize
    \centering
    \begin{tabular}{c|c|c|c}
    \hline
 \textbf{Data}   & \textbf{Set} & \textbf{$N$ of sentences} & \textbf{$N$ of words} \\\hline
 EWT & train & 12,543 & 204,579 \\
 & dev & 2,001 & 25,149 \\
 & test & 2,077 & 25,097 \\\hline
 Tweebank & train & 1,639 & 24,753 \\
 & dev & 710 & 11,742 \\
 & test & 1,201 & 19,112 \\\hline
 ESL & train & 4,124 & 78,541 \\
 & dev & 500 & 9,549 \\
 & test & 500 & 9,591 \\\hline
    \end{tabular}
    \caption{Descriptive statistics for out-of-domain data sets.}
    \label{tab:out_domain}
\end{table}

\begin{table}[h!]
\footnotesize
    \centering
    \begin{tabular}{c|c|c|c}
    \hline
 \textbf{Data} & \textbf{Parser} & \textbf{LM} & \textbf{LAS}  \\\hline
 EWT & MaChamp & \texttt{bert} & 90.49 \\
 & & \texttt{roberta} & 91.27 \\
 & & \texttt{twitter} & 90.84 \\\hline
  & Diaparser & \texttt{bert} & 89.89 \\
 & & \texttt{roberta} & 88.40 \\
 & & \texttt{twitter} & 88.26 \\\hline
 & UUParser & - & 82.68 \\\hline
 Tweebank & MaChamp & \texttt{bert} & 79.57 \\
 & & \texttt{roberta} & 80.49 \\
 & & \texttt{twitter} & 80.26 \\
 & Diaparser & \texttt{bert} & 80.01 \\
 & & \texttt{roberta} & 79.48 \\
 & & \texttt{twitter} & 79.65 \\\hline
 & UUParser & 69.54 \\\hline
  ESL & MaChamp & \texttt{bert} & 91.70 \\
 & & \texttt{roberta} & 92.27 \\
 & & \texttt{twitter} & 91.94 \\\hline
 & Diaparser & \texttt{bert} & 91.48 \\
 & & \texttt{roberta} & 90.74 \\
 & & \texttt{twitter} & 90.60 \\\hline
 & UUParser & - & 86.57 \\\hline
    \end{tabular}
    \caption{Parser evaluation results for out-of-domain data sets.}
    \label{tab:out_domain_eval}
\end{table}

We used three out-of-domain datasets of written English (Table~\ref{tab:out_domain}): (1) UD\_English-EWT (EWT)~\citep{silveira14gold}, which contains written texts from five genres
of web media, including weblogs, newsgroups, emails, reviews, and Yahoo! answers; (2) UD\_Tweebank (Tweebank)~\citep{liu-etal-2018-parsing}, which consists of English tweets; (3) UD\_English-ESL (ESL)~\citep{berzak2016tle}, which includes writing of learner English taken from the Cambridge Learner Corpus First Certificate in English dataset~\citep{yannakoudakis2011fce}.

For each of the aforementioned datasets, we trained the parsers described in Section~\ref{parsers} with their default parameters. We calculated micro unlabeled attachment scores and labeled attachment scores (LAS) to evaluate parser performance. Due to space limitation, throughout this paper we focus on reporting LAS. Parser evaluation results across three random seeds for all out-of-domain datasets are reported in Table~\ref{tab:out_domain_eval}. 

\subsection{Evaluation of out-of-domain parsers}

We first performed automatic part-of-speech tagging for all child and parent speech using the publicly-open NLP library Stanza~\citep{qi-etal-2020-stanza}. We then applied each of the out-of-domain parsers to child and parent speech within each age range of the child; parser performance was again index by LAS across 3 random seeds.
We foresee two directions regarding the parsing results.
On one hand, EWT has significantly much more data in contrast to Tweebank and ESL, which might lead to overall better results on child and parent speech.
That said, among the three out-of-domain datasets, the domains of Tweebank and ESL are possibly most relevant or more similar to child-parent interactions, in the sense that they are less ``formal" compared to EWT. This potentially means that parsers trained from Tweebank or ESL might outperform those based on EWT.

Note that to be fair for the out-of-domain parsers, here the evaluations did not consider new dependency relation subtypes that were introduced in our annotations, which are \textit{reparandum:restart} and \textit{reparandum:repetition}.

\begin{figure*}[h!]
     \centering
\begin{subfigure}{\textwidth}
  \centering
  \includegraphics[width=0.8\linewidth,height=2.5in]{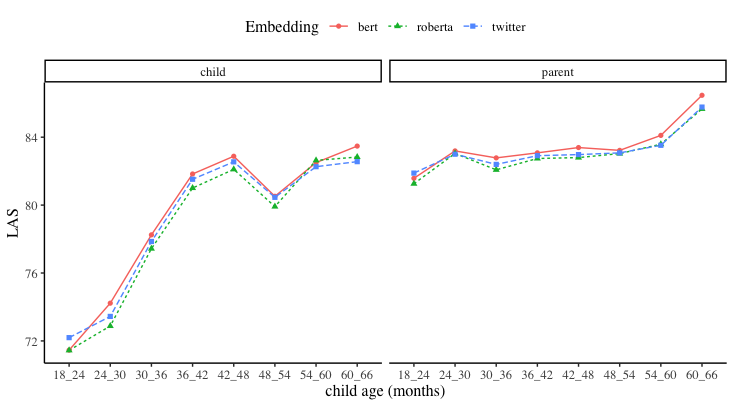}
  \caption{MaChamp}
  \label{fig:1}
\end{subfigure}
\begin{subfigure}{\textwidth}
  \centering
  \includegraphics[width=0.8\linewidth,height=2.5in]{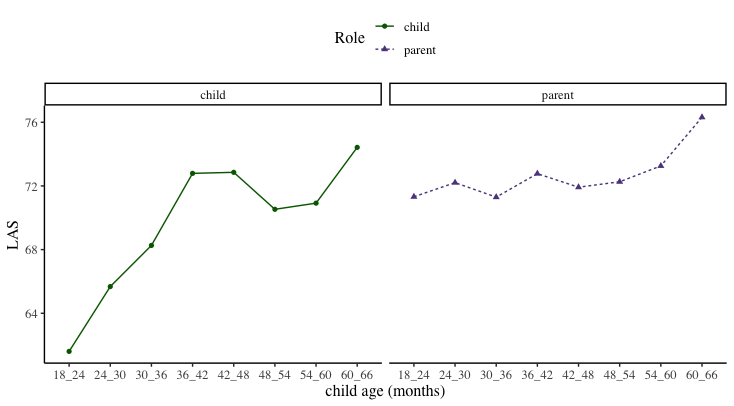}
  \caption{UUParser}
  \label{fig:2}
\end{subfigure}
         \caption{Evaluation of out-of-domain Tweebank parsers; at each age range of the children, the parser score for child (or parent) speech was averaged across all children (or parents) within that age range.}
        \label{fig:out_of_domain_eval}
\end{figure*}

\section{Results and analysis}
\label{results}

We present analysis mainly for results based on MaChamp; though we performed the same analysis for the other two parsers as well and there were no noticeable discrepancies in the patterns. (While there are differences in numerical results given that UUParser was not trained with neural LMs).

\subsection{Parent's speech}

When looking at parser performance for all parents across different age ranges of their children, 
on average parsers trained on the out-of-domain ESL dataset with \texttt{twitter} achieved the best result (83.88).
By contrast, the best parsers from EWT, which were also trained with \texttt{twitter} notably performed worse; the average difference in LAS ranges from 2.11 between the age range of 48-54 months, to 3.95 between the age range of 18 to 24.
On the other hand, the best parsers from Tweebank, trained with \texttt{bert}, achieved comparable performance (83.48) to the best parsers based on the ESL dataset.
What's worth noting here, is that comparing the parsing results of different LMs, the discrepancies in LAS for Tweebank are smaller than those for ESL.
For example, the biggest difference in LAS for Tweebank is 0.80 at the age range of 60-66 months between \texttt{bert} and \texttt{roberta};
whereas for ESL, the smallest discrepancy is 1.51 (48-54 months), and can be as large as 3.42 (18-24 months).
This means that parsers trained from Tweebank, which contains less than 1/8 of the amount of data from EWT, and around 1/3 from ESL, are able to lead to good and reliable performance.

So where do the discrepancies between parsers trained from different out-of-domain treebanks come from, especially during early ages of the children?
Comparing the best performing parsers based on EWT and those from Tweebank and ESL, it appears that 
for some utterances where the copula takes the form of `\textit{'s}', parsers trained from EWT errorneously annotated the copula as the subject, therefore assigning two subjects to the same syntactic head;
this accounts for around 17.07\% of all differences between the parsers' and our manual annotations.
This raises the worrisome question of why a structure where the syntactic head has two subjects would arise.
The most plausible answer is that such structures exist in the training set.
To check that, we looked into the training data of the EWT treebank and found five relevant sentences.
In these cases, the first subject between the two were incorrectly annotated as headed by the verb of the subordinate clause at a lower level 
(e.g., in \textit{\textbf{it} is not about how much \textbf{you} earn} (adapted), \textit{it} was annotated as the subject of \textit{earn}).
Similarly, in cases where `'s' has a dependency relation of \textit{aux}, the parsers trained from EWT also tended to 
assign it a relation of \textit{nsubj} instead; this led to around 7.69\% of all the discrepancies between parser and manual annotations.
The patterns described above seem to be the main and consistent explanation for the performance discrepancies between the best parsers trained from different out-of-domain treebanks, even when the children get older; between the age range of 54-60 months, around 11.30\% of all differences when comparing parsers's annotations to those done manually came from the parsers assigning \textit{nsubj} instead of \textit{cop}; and the number is approximately 11.99\% when the age range is between 60 and 66 months.

Now let us turn to the question of where do best-performing parsers trained from out-of-domain treebanks fall short in general? We note four cases here.
During early ages of the children, for parent speech, the dependency relation that results in the biggest discrepancy between parser and manual annotations 
is \textit{nmod:poss}, e.g., \textit{my book}, where the parsers annotated the relation as \textit{nmod}, which is less preferred in latest annotation guidelines of UD.
The second case that caused confusion for the parsers is when the sentences contain elements that should be annotated as \textit{discourse} and/or \textit{vocative} (e.g., \textit{hahaha I see}, \textit{Roman oh that is beautiful})
whereas the parsers appeared to more consistently annotate the first word in these instance as the root of the sentence. 

One other noteworthy example is when the parsers think of a \textit{vocative} as the subject  (e.g., \textit{Adam eat your soupe}) or sometimes object (e.g., \textit{sit Sarah}) of the utterance, since there is no punctuation in the annotations and the main clause of the sentence is subjectless or should not take an object at all.
The last one is \textit{conj}, which we used for cases of repetition or when the speaker appears to be listing individual nouns (e.g., \textit{orange grape apple}); the parsers, however, annotated the relation between pairs of nouns to be \textit{compound}, which was common in the training sets of the out-of-domain treebanks given that they were annotated (largely) based on UD guidelines.

As the children age, the aforementioned dependency relations still turned out to be the main reasons to explain parser performance, though to lesser extents. Note this was not because the parsers started correctly annotating these relations, but due to the fact that the relevant instances occur less frequently in parent speech at later age ranges of the children, at least in the data samples that we selected and annotated.
In addition, another dependency relation that the parsers tended to not annotate correctly at later developmental stages of the children is \textit{compound:prt}, in cases such as \textit{hurry up}; instead the parsers assigned \textit{compound}, which is not entirely wrong, but less precise.

\subsection{Child's speech}

Overall parser performance for child's speech demonstrates similar patterns to those found for parent's speech.
Across all age ranges, again the best parsers trained on the ESL treebank using \texttt{twitter} and the best parsers trained on the Tweebank with \texttt{bert} arrived at comparable performance (79.35 vs. 79.39); these parsers also outperformed the best parsers based on the EWT treebank (76.48), indicating that ESL and Tweebank, especially the latter, suffice for yielding reasonable parser performance despite their much smaller dataset scales in contrast to EWT.

In terms of what dependency relations resulted in the discrepancies between parser performance and manual annotations, we identified four main categories that are applicable especially during early ages of the children, which again are similar to those noted for parent speech.
The first one is \textit{conj}; the parsers tended to assign \textit{compound} for an utterance that consists mostly of individual nouns (\textit{book table pencil}) where none of the nouns modifies each other. 

The other three categories appear to be mainly caused by the utterances potentially having word omissions or ambiguous structures.
For instance, when the utterances consist of two words, where the first one was treated as the subject of the second (\textit{Adam home}) in our annotations, given the length of the utterances, the parsers again preferred to annotate \textit{Adam} as the \textit{compound} of \textit{home}.
Another case pertains \textit{nmod:poss}; in addition to assigning \textit{nmod} for utterances that contain possessives such as \textit{my}, the parsers also used \textit{nmod} for utterances that possibly involve the omission of the possisive marker `\textit{'s}'. 
The last one concerns the treatment of \textit{vocative}, which again sometimes were annotated by the parsers as \textit{nsubj} or \textit{obj}.


Notice there is small fluctuation in LAS between the age of 48-54 months. When looking at the differences between parser and manual annotations for this age range, we see an additional type of instance where the parsers committed errors, which involve a head verb with an adverbial modifier, such as \textit{bring them back}; rather than parsing \textit{back} as \textit{advmod} of \textit{bring}, the parsers assigned \textit{aux}.

In all, contrasting results for child speech to those for parent speech, not so surprisingly, parser performance is better overall for the latter. That said, when children reach later ages, the parsers' evaluation results ($\sim$87.83) approach those for parent speech ($\sim$89.13), suggesting that children's development of syntactic structures are becoming more parent-like.
While parser scores for parent speech slowly increase between the age range of 18-66 months, this progress is much more pronounced for child speech; in other words, we do see an overall improvement of parser performance as children progress along the syntactic developmental trajectory, particularly within 18-42 months.

\section{Ongoing in-domain parser training and evaluation}

After observing the performance of out-of-domain parsers, we now turn to evaluating the performance of parsers trained with in-domain data of child-parent interactions; in particular, these parsers are expected to pick up the new dependency relations for disfluencies that we introduced, at least to some extent.
Based on results from Section~\ref{results}, on average parsers trained with MaChamp using the \texttt{twitter} LM seemed to achieve the most stable performance, therefore we adopted the same parser setup here.

Our training scheme is as follows. Say we want to evaluate parser performance for Adam from the Brown Corpus. We first trained parsers using all the data from the other nine child-parent pairs; we then measured parser performance for both child and parent speech at each age range of the child using LAS averaged across 3 random seeds.



\section{Conclusion}

We present a wide-coverage dataset of child-parent interactions annotated with syntactic dependencies, along with detailed annotation guidelines extending the Universal Dependencies project. The dataset covers child and parent speech from the age range of 18-66 months for the children. Evaluations from graph-based and transition based dependency parsers with varying hyperparameters demonstrate that parsers trained using a relatively small amount of English tweets (Tweebank) are able to yield comparable or even outperform parsers trained from much larger dependency treebanks.
In addition, we observed the general trend that on average, parser performance increases as the children reach older ages, indicating that as children progress along their syntactic developmental trajectory, they start producing more cohesive structures but not too complex for the parsers to handle. We wait to verify this again with our ongoing work of in-domain parser training and evaluation..

\bibliography{tacl2021,custom}
\bibliographystyle{acl_natbib}





\end{document}

~\citep{gerdes-2009-speak,bazillon-2012-syntac}: spoken French; um and uh were not considered part of the dependency structure

~\citep{mikulova-2017-pdtsc} spoken Czech; consider disfluency as ungrammatical and reconstruct disfluency

~\citep{braggaar-2021-creat-univer} spoken Dutch-Frisian code-switched data

~\citep{odijk-2018-anncor-child-treeb} Dutch CHILDES data

~\citep{moore2015incremental} spoken learner English (focused on reparandum)

~\citep{caron-2019-surfac-syntac} Naija

~\citep{bechet-2014-adapt} phone-conversations of French

~\citep{davidson-2019-depen-parsin} spoken English dialogue human-machine conversation

~\citep{cetinoglu-2019-chall} Turkish-German code-switched conversations

~\citep{dobrovoljc-2018-er,dobrovoljc-2016-univer-depen} Slovenian

\begin{table*}[h!]
\footnotesize
\centering
\begin{tabular}{c|c|c|c|c|c|c|c|c}
\textbf{Child} & \textbf{18-24} & \textbf{24-30} & \textbf{30-36} & \textbf{36-42} & \textbf{42-48} & \textbf{48-54} & \textbf{54-60} & \textbf{60-66}  \\\hline
Violet & 1,730 & 1,855 & 1,889 & 1,886 & 1,894 &   &   \\
 & sents: 579 & sents: 449 & sents: 428 & sents: 451 & sents: 389 &   &   \\
 & MLU: 2.99 & MLU: 4.13 & MLU: 4.41 & MLU: 4.18 & MLU: 4.87 &   &   \\
Parent & 2,007 & 1,995 & 2,025 & 2,044 & 2,030 &   &   \\
 & sents: 316 & sents: 354 & sents: 305 & sents: 326 & sents: 235 &   &   \\
 & MLU: 6.35 & MLU: 5.64 & MLU: 6.64 & MLU: 6.27 & MLU: 8.64 &   &   \\\hline
Naima & 1,748 & 1,752 & 1,814 & 1,902 & 1,943 &   &   \\
 & sents: 503 & sents: 409 & sents: 369 & sents: 352 & sents: 284 &   &   \\
 & MLU: 3.48 & MLU: 4.28 & MLU: 4.92 & MLU: 5.4 & MLU: 6.84 &   &   \\
Parent & 2,019 & 2,007 & 2,005 & 2,032 & 2,019 &   &   \\
 & sents: 263 & sents: 271 & sents: 219 & sents: 312 & sents: 277 &   &   \\
 & MLU: 7.68 & MLU: 7.41 & MLU: 9.16 & MLU: 6.51 & MLU: 7.29 &   &   \\\hline
Thomas &   & 1,969 & 1,965 & 1,972 & 1,984 & 2,009 & 1,992 \\
 &   & sents: 703 & sents: 666 & sents: 469 & sents: 379 & sents: 380 & sents: 385 \\
 &   & MLU: 2.8 & MLU: 2.95 & MLU: 4.2 & MLU: 5.23 & MLU: 5.29 & MLU: 5.17 \\
Parent &   & 2,013 & 2,035 & 2,050 & 2,016 & 2,025 & 2,010 \\
 &   & sents: 348 & sents: 298 & sents: 293 & sents: 270 & sents: 297 & sents: 303 \\
 &   & MLU: 5.78 & MLU: 6.83 & MLU: 7.0 & MLU: 7.47 & MLU: 6.82 & MLU: 6.63 \\\hline
Adam &   & 2,012 & 2,004 & 2,016 & 2,021 & 2,012 & 2,026 & 2,025 \\
 &   & sents: 716 & sents: 608 & sents: 469 & sents: 443 & sents: 402 & sents: 376 & sents: 358 \\
 &   & MLU: 2.81 & MLU: 3.3 & MLU: 4.3 & MLU: 4.56 & MLU: 5.0 & MLU: 5.39 & MLU: 5.66 \\
Parent &   & 2,012 & 2,009 & 2,020 & 2,008 & 2,015 & 2,007 & 1,234 \\
 &   & sents: 385 & sents: 368 & sents: 358 & sents: 351 & sents: 350 & sents: 329 & sents: 204 \\
 &   & MLU: 5.23 & MLU: 5.46 & MLU: 5.64 & MLU: 5.72 & MLU: 5.76 & MLU: 6.1 & MLU: 6.05 \\\hline
Lily & 2,004 & 1,788 & 1,908 & 1,981 & 1,957 & 1,161 &   \\
 & sents: 641 & sents: 486 & sents: 394 & sents: 407 & sents: 337 & sents: 249 &   \\
 & MLU: 3.13 & MLU: 3.68 & MLU: 4.84 & MLU: 4.87 & MLU: 5.81 & MLU: 4.66 &   \\
Parent & 1,996 & 2,016 & 1,999 & 2,038 & 2,006 & 2,000 &   \\
 & sents: 309 & sents: 313 & sents: 298 & sents: 239 & sents: 270 & sents: 219 &   \\
 & MLU: 6.46 & MLU: 6.44 & MLU: 6.71 & MLU: 8.53 & MLU: 7.43 & MLU: 9.13 &   \\\hline
Laura & 1,930 & 1,925 & 1,955 & 1,912 & 1,956 & 1,943 & 1,954 \\
 & sents: 681 & sents: 529 & sents: 469 & sents: 418 & sents: 364 & sents: 379 & sents: 402 \\
 & MLU: 2.83 & MLU: 3.64 & MLU: 4.17 & MLU: 4.57 & MLU: 5.37 & MLU: 5.13 & MLU: 4.86 \\
Parent & 2,022 & 2,019 & 2,013 & 1,991 & 2,023 & 1,984 & 2,015 \\
 & sents: 415 & sents: 360 & sents: 358 & sents: 355 & sents: 321 & sents: 346 & sents: 325 \\
 & MLU: 4.87 & MLU: 5.61 & MLU: 5.62 & MLU: 5.61 & MLU: 6.3 & MLU: 5.73 & MLU: 6.2 \\\hline
Roman &   & 1,980 & 1,975 & 1,990 & 2,004 & 1,999 & 2,009 \\
 &   & sents: 493 & sents: 418 & sents: 355 & sents: 257 & sents: 298 & sents: 356 \\
 &   & MLU: 4.02 & MLU: 4.72 & MLU: 5.61 & MLU: 7.8 & MLU: 6.71 & MLU: 5.64 \\
Parent &   & 1,712 & 2,017 & 2,012 & 2,020 & 2,013 & 2,027 \\
 &   & sents: 278 & sents: 309 & sents: 321 & sents: 308 & sents: 287 & sents: 310 \\
 &   & MLU: 6.16 & MLU: 6.53 & MLU: 6.27 & MLU: 6.56 & MLU: 7.01 & MLU: 6.54 \\\hline
Emma &   &   & 2,007 & 2,014 & 2,029 & 2,025 &   \\
 &   &   & sents: 391 & sents: 349 & sents: 360 & sents: 381 &   \\
 &   &   & MLU: 5.13 & MLU: 5.77 & MLU: 5.64 & MLU: 5.31 &   \\
Parent &   &   & 2,020 & 2,038 & 2,015 & 1,455 &   \\
 &   &   & sents: 275 & sents: 280 & sents: 316 & sents: 236 &   \\
 &   &   & MLU: 7.35 & MLU: 7.28 & MLU: 6.38 & MLU: 6.17 &   \\\hline
Sarah &   & 2,014 & 2,008 & 2,011 & 2,007 & 2,020 & 2,015 & 2,028 \\
 &   & sents: 786 & sents: 715 & sents: 542 & sents: 439 & sents: 431 & sents: 392 & sents: 416 \\
 &   & MLU: 2.56 & MLU: 2.81 & MLU: 3.71 & MLU: 4.57 & MLU: 4.69 & MLU: 5.14 & MLU: 4.88 \\
Parent &   & 2,019 & 2,043 & 2,037 & 2,036 & 2,026 & 2,032 & 2,049 \\
 &   & sents: 381 & sents: 408 & sents: 384 & sents: 348 & sents: 339 & sents: 339 & sents: 318 \\
 &   & MLU: 5.3 & MLU: 5.01 & MLU: 5.3 & MLU: 5.85 & MLU: 5.98 & MLU: 5.99 & MLU: 6.44 \\\hline
Abe &   & 2,007 & 2,007 & 2,022 & 2,020 & 2,036 & 2,021 & 1,386 \\
 &   & sents: 471 & sents: 419 & sents: 328 & sents: 321 & sents: 303 & sents: 320 & sents: 234 \\
 &   & MLU: 4.26 & MLU: 4.79 & MLU: 6.16 & MLU: 6.29 & MLU: 6.72 & MLU: 6.32 & MLU: 5.92 \\
Parent &   & 2,028 & 2,040 & 2,025 & 2,020 & 2,017 & 2,019 \\
 &   & sents: 309 & sents: 359 & sents: 311 & sents: 364 & sents: 355 & sents: 351 \\
 &   & MLU: 6.56 & MLU: 5.68 & MLU: 6.51 & MLU: 5.55 & MLU: 5.68 & MLU: 5.75 \\\hline
\end{tabular}
\caption{Descriptive statistics for utterances of each child and their parent(s)}
\label{tab:descriptive_stats}
\end{table*}